\documentclass{article}
\usepackage{spconf,amsmath,graphicx}
\usepackage{amsmath,amsfonts,bm}
\usepackage{xcolor}
\usepackage{multirow}
\usepackage{booktabs}
\usepackage{url}
\usepackage{algorithm,algpseudocode}

\newcommand{\blind}{}

\title{Comparison of Soft and Hard Target RNN-T Distillation for Large-scale ASR}
\name{
\blind{
\begin{tabular}{c}
Dongseong Hwang, Khe Chai Sim, Yu Zhang, Trevor Strohman
\end{tabular}
}
}
\address{Google LLC, USA \\
\fontsize{9}{9}\selectfont\ttfamily\upshape
\{dongseong, khechai, ngyuzh, strohman\}@google.com}
\begin{document}
\ninept
\maketitle

% Example definitions.
% --------------------
\def\x{{\mathbf x}}
\def\L{{\cal L}}

\newcommand{\N}{\mathsf{N}}
\newcommand{\RDistBayes}{\hat{R}_*}
\newcommand{\REmp}{\hat{R}}

\newcommand{\gL}{\mathcal{L}}
\newcommand{\vx}{\bm{x}}
\newcommand{\vy}{\bm{y}}
\newcommand{\lm}{lm_u}
\newcommand{\am}{am_t}
\begin{abstract}
Knowledge distillation is an effective machine learning technique to transfer knowledge from a teacher model to a smaller student model, especially with unlabeled data.
In this paper, we focus on knowledge distillation for the RNN-T model, which is widely used in state-of-the-art (SoTA) automatic speech recognition (ASR).
Specifically, we compared using soft and hard target distillation to train large-scale RNN-T models on the LibriSpeech/LibriLight public dataset (60k hours) and our in-house data (600k hours). 
We found that hard targets are more effective when the teacher and student have different architecture, such as large teacher and small streaming student.
On the other hand, soft target distillation works better in self-training scenario like iterative large teacher training. 
For a large model with $0.6$B weights, we achieve a new SoTA word error rate (WER) on LibriSpeech (8\% relative improvement on dev-other) using Noisy Student Training with soft target distillation. It also allows our production teacher to adapt new data domain continuously.
\end{abstract}

\section{Introduction}
\label{sec:intro}
The success of end-to-end (E2E) speech recognition models \cite{sainath2020streaming,li2021better,narayanan2019recognizing} are highly dependent on having a large amount of high-quality transcribed speech data. However, it is expensive and difficult to get the high-quality human transcriptions, which restricts the development of automatic speech recognition (ASR). 

Knowledge distillation~\cite{hinton2015distilling} is an effective technique to transfer knowledge from a teacher model to a student model. Noisy Student Training (NST)~\cite{xie2020self,park2020improved} is a well established method that applies knowledge distillation in an iterative training fashion to progressively improve the model. For each iteration, NST transcribe a large amounts of unlabeled data from teacher model and use it to train the student model with data augmentation.
NST has been shown to be effectiveness for ImageNet~\cite{xie2020self} and the LibriSpeech automatic speech recognition task~\cite{park2020improved}.
%by ImageNet~\cite{imagenet2009} state-of-the-art (SoTA) results~\cite{xie2020self} and LibriSpeech~\cite{panayotov2015librispeech} SoTA results~\cite{park2020improved}.

In this paper, we explored knowledge distillation for the RNN-T~\cite{Graves2012} model. RNN-T is widely used in large-scale ASR systems~\cite{hwang2021large,zhang2022bigssl,hwang2022pseudo} and achieves state-of-the-art results on the LibriSpeech dataset~\cite{zhang2020pushing,xu2021self,chung2021w2v}. %However, RNN-T knowledge distillation method is not well established. 
NST training of RNN-T models was first studied in~\cite{park2020improved} using hard target distillation~\cite{hinton2015distilling,gou2021knowledge}, where the student model is trained using pseudo labels generated by a teacher model. Hard target distillation was also used in the follow-up works~\cite{zhang2020pushing,chung2021w2v} that further improved the SoTA results on LibriSpeech by combining with pre-training. More recently, soft target distillation for RNN-T was explored in~\cite{panchapagesan2021efficient, leal2021self}, where the KL divergence between the teacher and student output label distribution is used as the loss function, similar to those used in~\cite{hinton2015distilling}. 
However, it was only used for model compression~\cite{panchapagesan2021efficient} and streaming ASR models~\cite{leal2021self}.

%However, soft target distillation is not proven for LibriSpeech SoTA level.

Motivated by the success of using soft target distillation in image domain~\cite{xie2020self} and a recent theoretical analysis ~\cite{krishna2020distillation} claiming that soft target distillation is better than hard target, this paper investigates the optimal knowledge distillation method for the large-scale RNN-T architecture.
%In addition, the first NST paper~\cite{xie2020self} also uses soft target distillation on image domain. The recent theoretical analysis paper~\cite{krishna2020distillation} claims that soft target distillation is better than hard target, because the loss variance of soft target (i.e. Bayes-distilled risk $\RDistBayes( f; S )$) is smaller than the variance of hard target (i.e. Standard empirical risk $\REmp( f; S )$).
% $$ \mathbb{V}_{S \sim \Pr^{N}}\left[ \RDistBayes( f; S ) \right] \leq \mathbb{V}_{S \sim \Pr^{N}}\left[ \REmp( f; S ) \right], $$
%At the moment, hard target distillation is widely used for RNN-T distillation, while soft distillation is widely used in other domains. This paper investigates that the optimal knowledge distillation method for the SoTA level RNN-T architecture.

We demonstrate that soft target distillation brings better word error rate (WER) in self-training whose teacher and student have the same architecture. Otherwise, hard target distillation brings better WER. Teacher self-training with soft target distillation makes new LibriSpeech SoTA WER on a similar setup to W2v-BERT paper~\cite{chung2021w2v}; WERs (dev/dev-other/test/test-other) of Conformer 600M model without external language model from $1.3/2.6/1.4/2.7$ to $1.3/2.4/1.4/2.6$. In addition, we succeeded to train new production teacher without performance degradation by soft distillation, in the case that training data distribution is shifted.

The contributions of this work include the following: (1) \textbf{A systematic study of soft and hard target distillation on large-scale SoTA RNN-T.} (2) \textbf{A practical solution of soft/hard target distillation given different situation.} (3) \textbf{A more efficient way to do soft distillation and achieve new SoTA on LibriSpeech.}

\section{Related Work}
\label{sec:related}

\subsection{RNN-T model}
\label{sec:rnnt}

\begin{figure}[t]
\centering
\includegraphics[width=4.5cm]{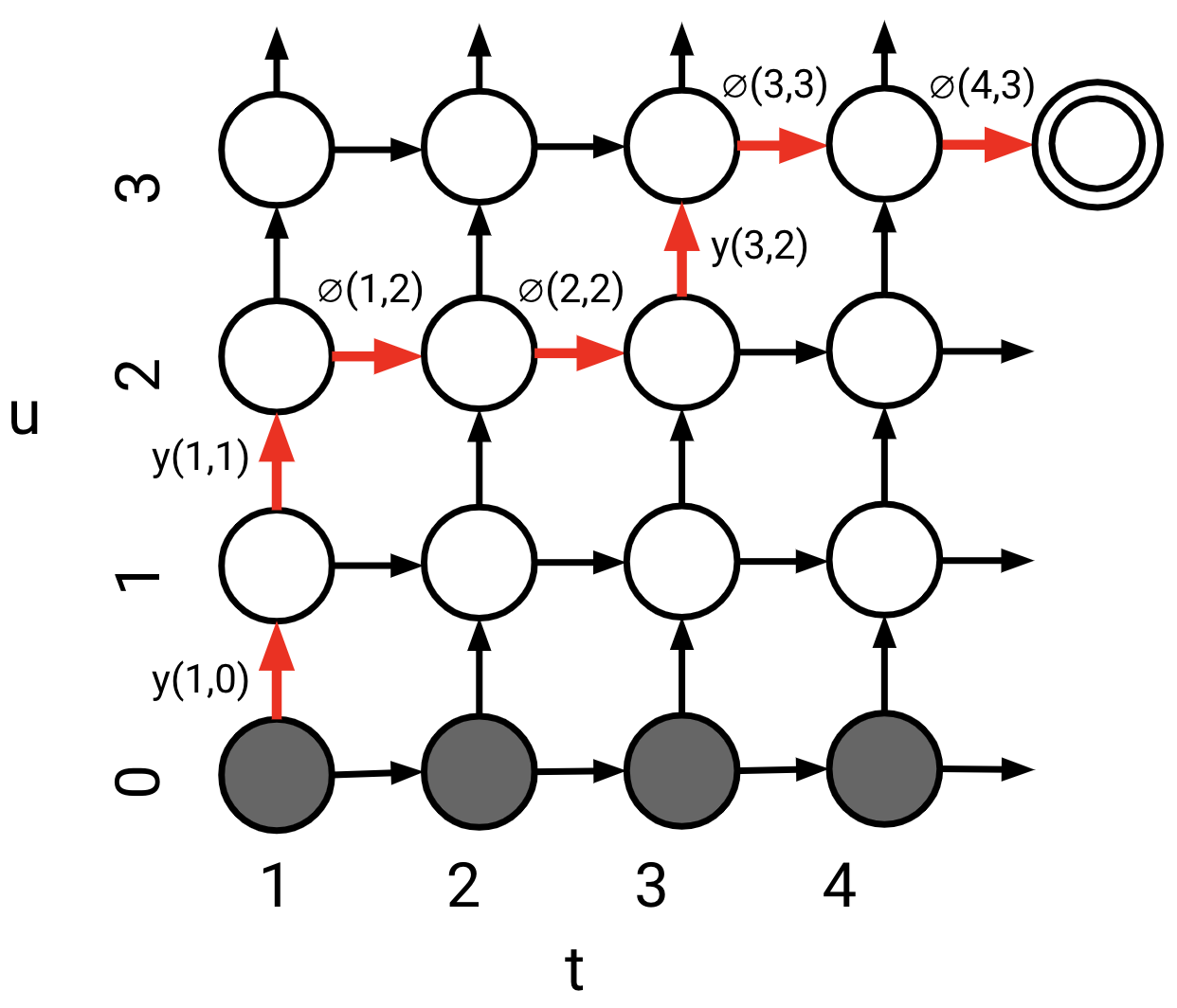}
\vspace{-0.1in}
\caption{RNN-T log probability $U \times T$ lattice, following \cite{Graves2012}.}
\label{fig:rnnt-lattice}
\vspace{-0.1in}
\end{figure}

In this section, we briefly summarize RNN Transducer~\cite{Graves2012} (RNN-T). The RNN-T loss is given by the sum of the negative log posterior probability for all possible sequences through the $U \times T$ lattice, as shown in Fig~\ref{fig:rnnt-lattice}, where $U$ is the length of target token sequence $\vy$ and $T$ is the number of audio input features $\vx$.
%$$\displaystyle \gL_\text{RNNT} = - \ln \bm{{P}}(\vy|\vx)$$
Each node at $(u,t)$ in the lattice represents a log probability made of a pair of acoustic ($\am$) and label ($\lm$) states. The RNN-T joint network combines these states to output the probabilities of the next label (such as characters or word pieces~\cite{schuster2012japanese}) token (vertical transition) or a special {\em blank} token (horizontal transition). The RNN-T loss can be computed efficiently using the forward-backward algorithm~\cite{bagby2018efficient}.
% RNN-T decoder produces language features $\lm$, and RNN-T encoder produces acoustic features $\am$. RNN-T joint layer produces the log probability $U \times T$ lattice, given $\lm$ and $\am$, as shown in Fig~\ref{fig:rnnt-lattice}.
% Each node at $(u,t)$ is the log probability of classes ($k$). The class is character or word piece~\cite{schuster2012japanese}. The each log probability is denoted as
% $${\ln \bm{{P}}(k|u,t)}$$

\subsection{RNN-T distillation}
\label{sec:rnnt-distil}

The first knowledge distillation~\cite{hinton2015distilling} paper introduces both soft target and hard target distillation in classification task. Soft target is a categorical distribution over all classes while hard target is a one-hot vector.

In RNN-T distillation, hard target is a transcript label, which is represented by a sequence of one-hot vectors. The first RNN-T NST~\cite{park2020improved} utilizes hard target distillation and then hard target distillation is widely used by follow-up LibriSpeech SoTA papers~\cite{zhang2020pushing,chung2021w2v,chiu2022self}.

As soft target is a categorical distribution, soft target distillation uses a KL divergence loss between teacher log probability and student log probability. The natural extension for RNN-T model is KL divergence over RNN-T log probability $U \times T$ lattice as shown in Eq~\ref{eq:grid_loss}. The recent RNN-T distillation papers~\cite{panchapagesan2021efficient, leal2021self} use this method.

\begin{align}
\displaystyle \gL_\text{KL} &= \sum_{u,t} \sum_k \bm{P_T}(k|u,t) \ln \frac{\bm{P_T}(k|u,t)}{\bm{P_S}(k|u,t)} \label{eq:grid_loss}
\end{align}

Soft RNN-T distillation is based on node-wise KL divergence which means that the student directly learns the alignment of the teacher, whereas for hard distillation, the student learns the alignment by RNN-T loss.

Efficient RNN-T distillation paper~\cite{panchapagesan2021efficient} proposes the memory efficient soft target distillation, because naive implementation requires $O(U \times T \times K)$ memory complexity in Eq~\ref{eq:grid_loss}. The efficient distillation distils 3 probabilities such as $y$, $blank$ and the remainder labels, instead of $k$ classes.

% https://docs.google.com/drawings/d/1_3-GjcZ2q37qEGSkS6XHm8nFOaE3DRWz5mUTI3Gv86A/edit
% \begin{figure}[htb]
%   \centering
%   \includegraphics[width=\linewidth]{nst.pdf}
%   \caption{Noisy student training (NST), following~\cite{hwang2022pseudo}}
%   \label{fig:nst}
% \end{figure}

\subsection{W2v-BERT}
\label{sec:w2v-bert}

Self-supervised learning techniques~\cite{chung2021w2v,chiu2022self,schneider2019wav2vec,baevski2020wav2vec,chen2021injecting, hsu2021hubert,baevski2022data2vec} have been shown to be effective for pre-training ASR models and achieved impressive performance for speech recognition tasks. We utilize self-supervised learning to pretrain the encoder of the student model before distillation. We compared two recent methods, W2v-BERT~\cite{chung2021w2v} and BEST-RQ~\cite{chiu2022self}, and found that the former achieved better WER results. Therefore, we use W2v-BERT pre-training for all the experiments in this paper.

\section{Method}
\label{sec:method}

\subsection{Multi-stage training using self/semi-supervised learning}
\label{sec:multi-stage}

We combine W2v-BERT and NST as follows:

1. Prepare the existing strong teacher.

2. Pretrain RNN-T encoder of the student by W2v-BERT.

3. Distillation from the teacher to the pre-trained student.

The existing strong teacher is bi-directional model trained by W2v-BERT and multi generation NST. We call the distillation target as student. In the paper, the student model is both large bi-directional model and small streaming model.

\subsection{Distillation methods}
\label{ssec:distillation_method}

We use a linear combination of the RNN-T loss ($\gL_\text{RNNT}$) and the KL-divergence loss ($\gL_\text{KL}$) as the overall training loss:
\begin{equation}
\gL = \alpha \gL_\text{RNNT} + (1 - \alpha) \gL_\text{KL}
\label{eq:overall_loss}
\end{equation}
%Eq~\ref{eq:overall_loss} shows our total distillation loss. 
It can express different training methods by adjusting $\alpha$. With $\alpha = 1$ and human labeled data, it is a conventional RNN-T training.
When pseudo-labeled data are used, we can achieve hard target distillation with $\alpha = 1$ and soft target distillation with $\alpha = 0$.
Furthermore, we can mix both the hard and soft target distillation by setting $\alpha$ to be between $0$ and $1$.
We can also mix soft target distillation with supervised training by using $\gL_\text{RNNT}$ on labeled data and $\gL_\text{KL}$ on unlabeled data, which is used in existing distillation work in other domains~\cite{hinton2015distilling,xie2020unsupervised,ridnik2022solving}. In this paper, most of the experiments use $\alpha = 0$ because we found that using $\gL_\text{KL}$ alone achieves better WERs, as shown in Section~\ref{sec:librispeech_results}.

We implement RNN-T soft target distillation efficiently. In Eq~\ref{eq:grid_loss}, the KL divergence for each $(u, t)$ can be computed independently, so it is free to compute all at once or break them into smaller groups to balance time and space trade-off. We found that computing $8$ time frames each iteration works best. With this implementation, vanilla soft target distillation and efficient RNN-T distillation~\cite{panchapagesan2021efficient} consume similar memory. See \ref{sec:disc_mode} for the experiments results.

\subsection{Spectrum Frequency augmentation}
\label{sec:freqaug}

Augmentation method choice is an important part in NST. 
We propose two different augmentation algorithms on top of SpecAugment~\cite{park2019specaugment}. These augmentation methods improve soft target distillation. See \ref{sec:diff_aug} for the experiment results.

\subsubsection{Frequency Warping}
\label{sec:freqwrap}
Frequency Warping warps a log mel spectrogram in the frequency axis, which imitates pitch shift. SpecAugment~\cite{park2019specaugment} introduces time warping. Frequency warping is the similar technique for the frequency axis as shown in Alg~\ref{alg:freq_warp}.

\begin{algorithm}[htb]
\caption{Frequency Warping}\label{alg:freq_warp}
\begin{algorithmic}
\Require input log mel feature, $\bm{x} \in {\rm I\!R}^{F \times T}$ $:$ $F$ is frequency dimension and $T$ is time sequence length.
\State 1. Draw $y$ in uniform random with range $(0,F)$ as the anchor point in Fig~\ref{fig:freq_warp}.
\State 2. Get the max distance $\Delta F = F \times \gamma_f$. $\gamma_f$ is frequency warping ratio, which is a hyper parameter.
\State 3. Draw $dy$ in uniform random with range $(-\Delta F, \Delta F)$.
\State 4. Get the destination point by $y' = y + dy$ clipped by range $(0,F)$.
\State 5. Warp the log mel feature from the anchor point $y$ to the destination point $y'$.
\end{algorithmic}
\end{algorithm}

% http://docs/drawings/d/1ARFhsu04XyaQqH4XZPnE0YmwqRtZzrfQhzHvLP0IlOQ?resourcekey=0-wVn71E0B0UN0EpyDEZ3A_g
\begin{figure}[htb]
  \centering
  \includegraphics[width=\linewidth]{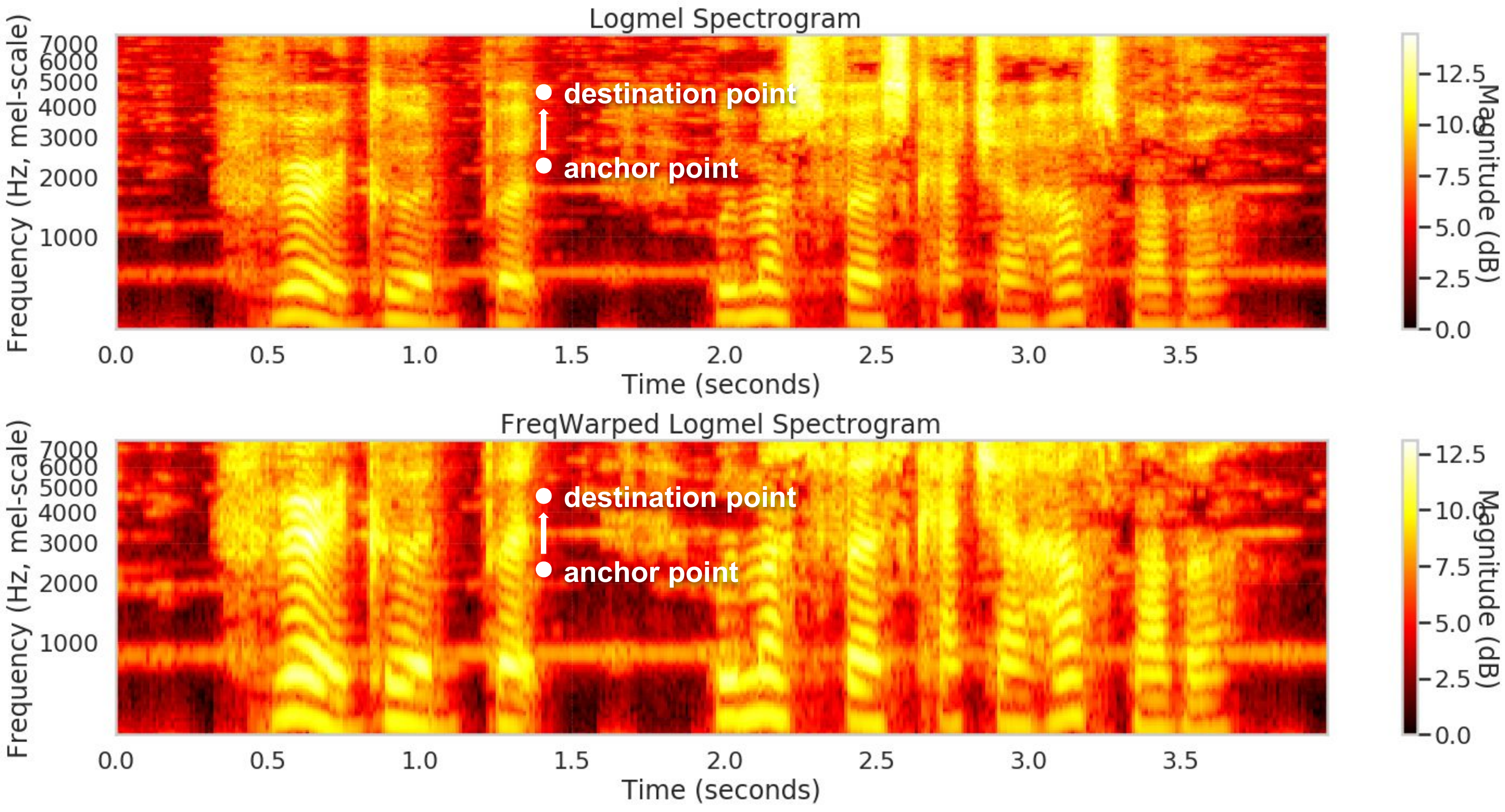}
  \caption{Frequency Warping converts the upper original log mel spectrogram to the lower augmented one.}
  \label{fig:freq_warp}
\end{figure}

\subsubsection{Frequency Noise}
\label{sec:freqnoise}
Frequency Noise is multiplicative noise to the log mel spectrogram, which imitates a random equalizer that boosts or lowers random frequencies. The Frequency Noise algorithms is shown in Alg~\ref{alg:freq_noise}.

\begin{algorithm}[htb]
\caption{Frequency Noise}\label{alg:freq_noise}
\begin{algorithmic}
\Require input log mel feature, $\bm{x} \in {\rm I\!R}^{F \times T}$ $:$ $F$ is frequency dimension and $T$ is time sequence length.
\State 1. Draw $\sigma_{f}$ in uniform random with range $(0,\sigma_{noise})$. $\sigma_{noise}$ is a hyper parameter.
\State 2. Draw $\bm{f_{noise}} \in {\rm I\!R}^{F}$ in Gaussian distribution $\mathcal{N}(1,\,\sigma_{f})$.
\State 3. Multiply $\bm{f_{noise}} \times \bm{x}$.
% \State 3. Add $\ln{\bm{f_{noise}}} + \bm{x}$ because $\bm{x}$ is in log domain.
\end{algorithmic}
\end{algorithm}

% Multiplication in frequency domain is convolution in time domain. Frequency Noise algorithm is equivalent to convolution operation with a periodic random kernel in time domain. The periodic random kernel is $\bm{f_{noise}}$ in time domain. Convolution operation of a periodic random kernel~\cite{whitenoise} represents a signal filtered by white noise. So Frequency Noise algorithm imitates a signal in white noise environment.

\subsection{Enhanced Conformer}
\label{ssec:conformer_v2}

The Conformer~\cite{conf} is widely used for state-of-the-art ASR systems. On the other hand, Primer~\cite{so2021searching} is the SoTA Transformer~\cite{vaswani2017attention} architecture for autoregressive language modeling, as discovered by AutoML. It was found to reduce the training costs by 4x and achieved better performance~\cite{so2021searching}.

Motivated by the use of the Squared ReLU activation function in Primer, we change the activation function for feed forward modules in Conformer from SWISH to Squared ReLU and found it to be beneficial, as shown in the experiment results in Section~\ref{sec:baseline}.

Primer also introduces depth separable convolution for $Q, K, V$. However, we did not find it useful for our setup. 
%Not all lessons of Primer are applicable to Conformer mainly because Primer is designed for decoder while Conformer is designed for encoder. 
In this paper, we use Conformer with Squared ReLU in all the experiments.

\section{Experiment}
\label{sec:exp}

\subsection{Experiment Setup}
\label{sec:experiment_setup}

\subsubsection{Model Architecture}
\label{sec:model_architecture}
We consider two types of models, one for the bi-directional teacher model and another for the bi-directional/streaming student model. 
Both models use $80$-dimension log mel features \cite{narayanan2019recognizing} as input features. The outputs correspond to $1,024$ word pieces~\cite{schuster2012japanese}.

The architecture and training procedure of the teacher model is similar to Conformer XL model ($0.6$B) in~\cite{zhang2020pushing}. The audio encoder has $24$ Conformer blocks~\cite{conf} with model dimension $1024$. The convolution kernel size is $5$ and the self-attention layer consists of $8$ heads with $128$ left and right context length. The decoder consists of a 2-layer LSTM~\cite{hochreiter1997long} label encoder with $2048$ units projected down to $640$ output units, and a joint network with a single feed-forward layer with $640$ units. The total number of weights is $0.6$B. The model is trained by minimizing the RNN-T loss~\cite{Graves2012}. 

The streaming student model is a smaller version of the teacher model. The audio encoder has $17$ Conformer blocks with model dimension $512$. We prepare both bi-directional and streaming students. Streaming student uses causal Conformer blocks. This model has a total of number $150$M weights.

\subsubsection{Data}
\label{ssec:data}

We conduct our experiments on both the public data and in-house data. First, public data consists of LibriSpeech and LibriLight dataset as shown in Tab~\ref{tab:librispeech}. \texttt{LL} is the main target of RNN-T distillation experiments as \texttt{LL} does not have labels. The teacher model produces pseudo labels for hard target distillation. The teacher model produces RNN-T $U \times T$ log probability lattice for soft distillation.

\begin{table}[htb]
    \centering
    \caption{Overview of LibriSpeech datasets.}
    \label{tab:librispeech}
    \begin{tabular}{ccc}
        \toprule
        \textbf{Data set} & \textbf{Label} & \textbf{Hours} \\
        \midrule
         LibriSpeech (\texttt{LS}) & human & $960$ \\
         LibriLight (\texttt{LL}) & unlabel & $60$k \\
        \bottomrule
    \end{tabular}
    \vspace{-0.4cm}
\end{table}

Second, we use a large multi-domain (\texttt{MD}) English dataset~\cite{Narayanan2018} as in-house dataset. The data consists of utterances from multiple domains, such as voice search (\texttt{VS}), medium-form (\texttt{MF}) and YouTube (\texttt{YT}). All the data are anonymized and hand-transcribed \blind{following Google AI principles~\cite{googleai}}. \texttt{VS} is mostly voice command. \texttt{MF} is mostly natural conversation. The \texttt{YT} labels are generated from YouTube video transcription with confidence filter~\cite{liao2013large}.

\begin{table}[htb]
    \centering
    \caption{Overview of in-house datasets. \texttt{MD} denotes all the supervised data (\texttt{VS} + \texttt{MF} + \texttt{YT}, $575$k hrs). \texttt{MD\textsubscript{new}} denotes the new data (\texttt{VS\textsubscript{unsup}} + \texttt{MF\textsubscript{unsup} + \texttt{YT}}, $615$k hrs).}
    \label{tab:data}
    \begin{tabular}{ccc}
        \toprule
        \textbf{Data set} & \textbf{Label} & \textbf{Hours} \\
        \midrule
         Voice search (\texttt{VS}) & human & $27$k \\
         Medium-form (\texttt{MF}) & human & $26$k \\
         Youtube (\texttt{YT}) & semi & $440$k \\
        \midrule
         Voice search (\texttt{VS\textsubscript{unsup}}) & unlabel & $150$k \\
         Medium-form (\texttt{MF\textsubscript{unsup}}) & unlabel & $25$k \\
        \bottomrule
    \end{tabular}
    \vspace{-0.4cm}
\end{table}

\subsection{Prepare the baseline teacher}
\label{sec:baseline}

We enhance RNN-T conformer model using Squared ReLU activation as proposed in \ref{ssec:conformer_v2}. It makes clear improvement compared to Conformer XL model of W2v-BERT paper~\cite{chung2021w2v}. As shown in Tab~\ref{tab:primer}, Squared ReLU experiment has better WERs than W2v-BERT XL (600M parameters) model~\cite{chung2021w2v}, which uses SWISH activation. \texttt{B0} is trained with the same methodology as W2v-BERT XL model~\cite{chung2021w2v}, such as W2v-BERT pre-train and then NST with hard target pseudo labels. \texttt{B0} goes through multiple NST generations until WERs are converged, which is 4 generations. The rest of experiments use Conformer with Squared ReLU activation, and \texttt{B0} as the teacher.

\begin{table}[htb]
    \centering
    \caption{LibriSpeech WERs comparison for the enhanced baseline.} 
    \label{tab:primer}
    \begin{tabular}{lcccc}
        \toprule
        {\multirow{2}*{\textbf{Model}}}& \multicolumn{4}{c}{\textbf{WER}} \\
         & {\textbf{dev}} & {\textbf{dev-other}} & {\textbf{test}} & {\textbf{test-other}} \\
        \midrule
        {W2v-BERT XL~\cite{chung2021w2v}} & ${1.3}$  & ${2.6}$ & ${1.4}$  & ${2.7}$ \\
        {B0 (Squared ReLU)} & ${1.3}$  & $\textbf{2.5}$ & ${1.4}$  & $\textbf{2.6}$ \\
        \bottomrule
    \end{tabular}
    %\vspace{-0.4cm}
\end{table}

\subsection{LibriSpeech results}
\label{sec:librispeech_results}

\subsubsection{Large bi-directional self-training}
\label{ssec:teacher}

We distil the baseline teacher (\texttt{B0}) to the same architecture student model. As shown in Tab~\ref{tab:teacher}, soft target distillation (\texttt{E1}) makes better WERs in self-training scenario. Hard target distillation (\texttt{E0}) has slightly worse WERs than the teacher (\texttt{B0}).

All LibriSpeech near-SoTA papers in Tab~\ref{tab:teacher} use hard target distillation. W2v-BERT~\cite{chung2021w2v} has LibriSpeech SoTA WERs before. WERs of soft target distillation (E1) is better, especially dev-other and test-other. \textbf{We have new SoTA WERs for 600M parameters models}.

\begin{table}[htb]
    \centering
    \caption{LibriSpeech WERs comparison for teacher models, without language model (LM). XL denotes 600M model and XXL denotes 1B model.} 
    \label{tab:teacher}
    \begin{tabular}{lcccc}
        \toprule
        {\multirow{2}*{\textbf{Model}}}& \multicolumn{4}{c}{\textbf{WER}} \\
         & {\textbf{dev}} & {\textbf{dev-other}} & {\textbf{test}} & {\textbf{test-other}} \\
        \midrule
        {B0 (teacher XL)} & ${1.3}$  & ${2.5}$ & ${1.4}$  & ${2.6}$ \\
        \midrule
        {E0 (hard target, XL)} & ${1.3}$  & ${2.5}$ & ${1.4}$  & ${2.7}$ \\
        {E1 (soft target, XL)} & ${1.3}$  & $\textbf{2.4}$ & ${1.4}$  & $\textbf{2.6}$ \\
        \midrule
        {BEST-RQ XL~\cite{chiu2022self}} & ${1.5}$  & ${2.8}$ & ${1.6}$  & ${2.9}$ \\
        {W2v-BERT XL~\cite{chung2021w2v}} & ${1.3}$  & ${2.6}$ & ${1.4}$  & ${2.7}$ \\
        \midrule
        {wav2vec2 XXL~\cite{zhang2020pushing}} & ${1.3}$  & ${2.7}$ & ${1.5}$  & ${2.7}$ \\
        {W2v-BERT XXL~\cite{chung2021w2v}} & ${1.4}$  & $\textbf{2.4}$ & ${1.4}$  & $\textbf{2.5}$ \\
        \bottomrule
    \end{tabular}
    %\vspace{-0.4cm}
\end{table}

\subsubsection{Small bi-directional student model}
\label{ssec:student}

As shown in Tab~\ref{tab:student}, hard target distillation makes better WERs than others, for 150M parameters student model. Soft target distillation still makes better WERs than RNN-T baseline (B1).
Soft target distillation does not work well when the architecture of teacher and student is different. On the other hand, hard target is architecture agnostic as hard target is just pseudo label.

\texttt{E4} combines hard target and soft target distillation by setting $\alpha = 0.5$ in Eq~\ref{eq:overall_loss}. The results are as bad as soft target distillation (\texttt{E3}). Soft target distillation loss (i.e. the KL term in Eq~\ref{eq:overall_loss}) with mismatching teacher and student logits affects the training badly, although the loss includes RNN-T loss. We assume there is an alignment mismatch between RNN-T loss and KL divergence loss, which affects the training badly.

\begin{table}[htb]
    \centering
    \caption{LibriSpeech WERs comparison for 150M student models.} 
    \label{tab:student}
    \begin{tabular}{lcccc}
        \toprule
        {\multirow{2}*{\textbf{Model}}}& \multicolumn{4}{c}{\textbf{WER}} \\
         & {\textbf{dev}} & {\textbf{dev-other}} & {\textbf{test}} & {\textbf{test-other}} \\
        \midrule
        {B0 (teacher XL)} & ${1.3}$  & ${2.5}$ & ${1.4}$  & ${2.6}$ \\
        {B1 (student)} & ${1.9}$  & ${4.4}$ & ${2.1}$  & ${4.6}$ \\
        \midrule
        {E2 (hard target)} & $\textbf{1.5}$  & $\textbf{3.6}$ & $\textbf{1.7}$  & $\textbf{3.6}$ \\
        {E3 (soft target)} & ${1.7}$  & ${3.7}$ & ${1.8}$  & ${3.8}$ \\
        {E4 (hard+soft target)} & ${1.7}$  & ${3.7}$ & ${1.8}$  & ${3.8}$ \\
        \bottomrule
    \end{tabular}
    %\vspace{-0.4cm}
\end{table}

\subsubsection{Small streaming student model}
\label{ssec:streaming}

Soft target distillation is even worse with a streaming student because the architecture is more different. As shown in Tab~\ref{tab:streaming}, hard target distillation (\texttt{E5}) makes the best WERs for 150M parameters streaming student model. Soft target distillation (\texttt{E6}) makes even worse WERs than RNN-T baseline (\texttt{B2}). 
\texttt{E7} combines hard target and soft target distillation, which has the worse WERs among all the streaming experiments. Soft target distillation loss (i.e. KL divergence in Eq~\ref{eq:overall_loss}) with mismatching teacher and student logits fights against hard target distillation loss (i.e. RNN-T loss in Eq~\ref{eq:overall_loss}), which results in bad WERs.

\begin{table}[htb]
    \centering
    \caption{LibriSpeech WERs comparison for streaming 150M student models.} 
    \label{tab:streaming}
    \begin{tabular}{lcccc}
        \toprule
        {\multirow{2}*{\textbf{Model}}}& \multicolumn{4}{c}{\textbf{WER}} \\
         & {\textbf{dev}} & {\textbf{dev-other}} & {\textbf{test}} & {\textbf{test-other}} \\
        \midrule
        {B0 (teacher XL)} & ${1.3}$  & ${2.5}$ & ${1.4}$  & ${2.6}$ \\
        {B2 (student)} & ${4.2}$  & ${10.4}$ & ${4.8}$  & ${9.5}$ \\
        \midrule
        {E5 (hard target)} & $\textbf{4.0}$  & $\textbf{9.4}$ & $\textbf{4.4}$  & $\textbf{8.6}$ \\
        {E6 (soft target)} & ${4.4}$  & ${11.1}$ & ${4.6}$  & ${11.1}$ \\
        {E7 (hard+soft target)} & ${4.6}$  & ${11.8}$ & ${4.7}$  & ${11.7}$ \\
        \bottomrule
    \end{tabular}
    % \vspace{-0.4cm}
\end{table}

\subsection{Production data experiments}
\label{ssec:prod_results}

We compare hard target and soft target distillation with our in-house data as mentioned in Sec~\ref{ssec:data}. We conduct distillation experiments with the existing strong teacher from \blind{the previous work~\cite{hwang2022pseudo}}. The existing teacher is \texttt{B3} in Tab~\ref{tab:prod} which is trained by W2v-BERT and multi generations NST with \texttt{MD} data as shown in Tab~\ref{tab:data}. The model architecture is very similar to LibriSpeech Conformer XL teacher. The biggest difference is that the model input feature is $4$ contiguous frames of $128$-dimension log-mel features \cite{narayanan2019recognizing} sub-sampled by a factor of $3$ and a one-hot domain-id vector of size $16$.

First, we conduct both hard and soft target distillation on the same architecture student model with the same \texttt{MD} data. As shown in the second group of Tab~\ref{tab:prod}, soft target distillation (\texttt{E9}) maintains the WERs of the teacher (\texttt{B3}), but hard target distillation (\texttt{E8}) degrades the performance. \texttt{B3} from the previous work~\cite{hwang2022pseudo} uses hard target distillation but has better WERs than \texttt{E8}, because it mixes pseudo labels with $1k$ hours human labels. Soft target distillation successfully maintains teacher WERs without any human labels. 

Second, we conduct the distillation experiments with newer and bigger data (\texttt{MD\textsubscript{new}}). As shown in the third group of Tab~\ref{tab:prod}, soft target distillation (\texttt{E11}) improves WERs, while hard target distillation (\texttt{E11}) degrades WERs. Soft target distillation is more robust in data domain shift scenario. \textbf{It enables the teacher to adapt new data without human label. It allows our production teacher to adapt new data domain in continual learning}.

\begin{table}[htb]
    \centering
    \caption{In-house data WERs comparison for production 600M teacher models. \texttt{E8} and \texttt{E9} are trained with the same data to the teacher \texttt{B3}. \texttt{E10} and \texttt{E11} are trained with out of domain data. Soft distillation (\texttt{E11}) recovers in-domain \texttt{VS} and \texttt{MF} WERs without in-domain training data.} 
    \label{tab:prod}
    \begin{tabular}{lcccc}
        \toprule
        {\multirow{2}*{\textbf{Model}}} & \multirow{2}*{\textbf{Data}} & \multicolumn{3}{c}{\textbf{WER ($\%$)}} \\
         & {} & {\textbf{VS}} & {\textbf{MF}} & {\textbf{YT}} \\
        \midrule
        {B3 (old teacher)} & \texttt{MD} & ${4.1}$ & ${4.3}$ & ${8.0}$ \\
        \midrule
        {E8 (hard target)} & \texttt{MD} & ${4.2}$ & ${4.4}$ & ${8.1}$ \\
        {E9 (soft target)} & \texttt{MD} & $\textbf{4.1}$ & $\textbf{4.3}$ & ${8.0}$ \\
        \midrule
        {E10 (hard target)} & \texttt{MD\textsubscript{new}} & ${4.3}$ & ${4.4}$ & ${7.9}$ \\
        {E11 (soft target)} & \texttt{MD\textsubscript{new}} & $\textbf{4.1}$ & $\textbf{4.3}$ & $\textbf{7.8}$ \\
        \bottomrule
    \end{tabular}
    %\vspace{-0.4cm}
\end{table}

\section{Discussion}
\label{sec:disc}

\subsection{Soft target vs hard target}
\label{sec:subsubhead}

As shown in Sec~\ref{sec:librispeech_results} and Sec~\ref{ssec:prod_results}, soft target distillation works better in self-training, but hard target distillation works better when the teacher and student have different architecture. Hard targets allow the student model to learn the alignment by itself using RNN-T loss and therefore works well when the teacher and student have different architecture such as bi-directional teacher and streaming student. On the other hand, soft distillation computes node-wise KL divergence over the $U \times T$ lattice, so it assumes that both the teacher and student models have similar optimal alignment. This is true when the teacher and student have the same architecture. When the alignment is not an issue, soft targets can transfer more information. We remain making alignment aware soft target distillation for future study. 

\subsection{Temperature scaling}
\label{sec:temperature_scaling}

Temperature scaling is widely used for distillation learning. Teacher logits are scaled by temperature $\tau$. High $\tau$ has smoothing effect, while low $\tau$ has sharpening effect. For example, Hinton's label smoothing paper~\cite{muller2019does} uses teacher $\tau = 1.9$, while UDA~\cite{xie2020unsupervised} paper uses teacher $\tau = 0.4$. The recent ImageNet SoTA distillation paper~\cite{ridnik2022solving} claims $\tau = 1$ works best.
As shown in Tab~\ref{tab:temp}, RNNT distillation also works best with $\tau = 1$. Interestingly, low teacher $\tau$ or high student $\tau$ works more stable.

\begin{table}[htb]
    \centering
    \caption{Soft distillation LibriSpeech WERs comparison for different $\tau$.} 
    \label{tab:temp}
    \begin{tabular}{lcccc}
        \toprule
        {\multirow{2}*{\textbf{Model}}}& \multicolumn{4}{c}{\textbf{WER}} \\
         & {\textbf{dev}} & {\textbf{dev-other}} & {\textbf{test}} & {\textbf{test-other}} \\
        \midrule
        {teacher $\tau=0.1$} & ${1.4}$  & ${2.5}$ & ${1.4}$  & ${2.7}$ \\
        {teacher $\tau=0.5$} & ${1.3}$  & ${2.6}$ & ${1.4}$  & ${2.7}$ \\
        {E12 (teacher $\tau=1$)} & $\textbf{1.3}$  & $\textbf{2.5}$ & $\textbf{1.4}$  & $\textbf{2.6}$ \\
        {teacher $\tau=2$} & ${1.6}$  & ${2.9}$ & ${1.7}$  & ${3.0}$ \\
        \midrule
        {student $\tau=0.5$} & ${2.3}$  & ${4.5}$ & ${2.6}$  & ${4.6}$ \\
        {E12 (student $\tau=1$)} & $\textbf{1.3}$  & $\textbf{2.5}$ & $\textbf{1.4}$  & $\textbf{2.6}$ \\
        {student $\tau=1.5$} & ${1.3}$  & ${2.6}$ & ${1.4}$  & ${2.6}$ \\
        \bottomrule
    \end{tabular}
    %\vspace{-0.4cm}
\end{table}

\subsection{Self-sup pretrain for student}
\label{sec:self-sup}

So far, we pretrain the student using W2v-BERT before distillation. When the student is randomly initialized, soft target distillation (\texttt{E13}) could not make the optimal WERs, as shown in Tab~\ref{tab:self-sup}. All distillation experiments has better WERs than RNN-T supervised train with only LibriSpeech data (\texttt{B4}). It demonstrates $1k$ hours speech data is not enough to train Conformer RNN-T model, and RNN-T distillation is a great solution.
% Soft target distillation with random initialization (\texttt{E13}) is better than hard target counter-part.
\begin{table}[htb]
    \centering
    \caption{Soft distillation LibriSpeech WERs comparison for different student initialization.} 
    \label{tab:self-sup}
    \begin{tabular}{lcccc}
        \toprule
        {\multirow{2}*{\textbf{Model}}}& \multicolumn{4}{c}{\textbf{WER}} \\
         & {\textbf{dev}} & {\textbf{dev-other}} & {\textbf{test}} & {\textbf{test-other}} \\
        \midrule
        {B4 (LibriSpeech)} & ${2.5}$  & ${6.1}$ & ${2.7}$  & ${6.2}$ \\
        \midrule
        {E13 (Random init)} & ${1.4}$  & ${3.0}$ & ${1.5}$  & ${3.1}$ \\
        {E12 (W2v-BERT)} & $\textbf{1.3}$  & $\textbf{2.5}$ & $\textbf{1.4}$  & $\textbf{2.6}$ \\
        % \midrule
        % {E13 + hard target} & ${1.4}$  & ${3.2}$ & ${1.6}$  & ${3.3}$ \\
        \bottomrule
    \end{tabular}
    %\vspace{-0.4cm}
\end{table}

\subsection{Frequency augmentation}
\label{sec:diff_aug}

Augmentation is key element in NST as the original paper~\cite{xie2020self} argued. Without augmentation, the student is merely distilling the knowledge of the teacher, which results in the same or worse performance. With augmentation, the student outperforms the teacher because it encourages the student to produce robust outputs on noisy input features. We investigate what augmentation the noisy student need in speech domain.

The start point is SpecAugment~\cite{park2019specaugment}. We use the same hyper parameters to the ASR NST paper~\cite{park2020improved}. We use two frequency masks with max mask size ($F$) $27$, and ten time masks with max mask size ($T$) $40$.
Interestingly, soft target distillation without any augmentations is as good as SpecAugment (\texttt{E12}), unlike NST papers reported, as shown in Tab~\ref{tab:disc_aug}. When applying SpecAugment to both teacher and student inputs, the WERs are bad.

\ref{sec:freqaug} introduces FreqWarp and FreqNoise. FreqAug denotes the combination of FreqWarp and FreqNoise. We search the optimal hyper parameters for both augmentations, which are that the frequency warping ratio ($\gamma_f$) is $75\%$ and the frequency noise stddev ($\sigma_{noise}$) is $0.14$. Soft target distillation with FreqAug (\texttt{E1}) has better WERs than any other augmentation experiments. However, hard target distillation with FreqAug has the same WERs as one with SpecAugment.

SpecAugment is random masking-out technique, which encourages the student to guess logits for missing features. FreqAug is more natural augmentation. FreqWarp mimics pitch shift and FreqNoise mimics noisy environments. FreqAug encourages the student to produce the similar logits in the natural noisy situation, which results in more generalized ASR model. 

\begin{table}[htb]
    \centering
    \caption{Soft distillation LibriSpeech WERs comparison for different augmentations.} 
    \label{tab:disc_aug}
    \begin{tabular}{lcccc}
        \toprule
        {\multirow{2}*{\textbf{Augmentation}}}& \multicolumn{4}{c}{\textbf{WER}} \\
         & {\textbf{dev}} & {\textbf{dev-other}} & {\textbf{test}} & {\textbf{test-other}} \\
        \midrule
        {No Augment} & ${1.3}$  & ${2.5}$ & ${1.4}$  & ${2.6}$ \\
        \midrule
        {E12 (SpecAug)} & ${1.3}$  & ${2.5}$ & ${1.4}$  & ${2.6}$ \\
        {E1 (FreqAug)} & $\textbf{1.3}$  & $\textbf{2.4}$ & $\textbf{1.4}$  & $\textbf{2.6}$ \\
        {E12 + FreqAug} & ${1.3}$  & ${2.5}$ & ${1.4}$  & ${2.7}$ \\
        \midrule
        {Noisy teacher} & ${1.5}$  & ${2.8}$ & ${1.6}$  & ${3.0}$ \\
        \bottomrule
    \end{tabular}
    %\vspace{-0.4cm}
\end{table}

\subsection{Different soft target distillation comparison}
\label{sec:disc_mode}

As shown in Tab~\ref{tab:disc_mode}, vanilla soft target distillation (\texttt{E12}) is better than efficient RNN-T distillation~\cite{panchapagesan2021efficient}, which is an approximation for speed and memory. Efficient distillation does not save additional memory. E12 consumes 13.7GB and efficient distillation consumes 13.8GB in TPUv3~\cite{jouppi2017datacenter}. The training speed is also similar.
% Soft distillation with the $U \times T$ lattice weighted by teacher RNN-T probability does not work well.

\begin{table}[htb]
    \centering
    \caption{LibriSpeech WERs comparison bewteen vanilla RNN-T distillation and efficient distillation~\cite{panchapagesan2021efficient}.} 
    \label{tab:disc_mode}
    \begin{tabular}{lcccc}
        \toprule
        {\multirow{2}*{\textbf{Model}}}& \multicolumn{4}{c}{\textbf{WER}} \\
         & {\textbf{dev}} & {\textbf{dev-other}} & {\textbf{test}} & {\textbf{test-other}} \\
        \midrule
        {E12 (soft target)} & $\textbf{1.3}$  & $\textbf{2.5}$ & $\textbf{1.4}$  & $\textbf{2.6}$ \\
        {Efficient~\cite{panchapagesan2021efficient}} & ${1.4}$  & ${2.5}$ & ${1.5}$  & ${2.6}$ \\
        % {Weighted} & ${5.8}$  & ${7.7}$ & ${5.7}$  & ${8.0}$ \\
        \bottomrule
    \end{tabular}
    %\vspace{-0.4cm}
\end{table}

\subsection{Limitation}
\label{sec:limit}

In this work, we explore optimal distillation method given the existing strong teacher. In the case that we do not have strong teacher, the suggested method may not work without modification. Self-Adaptive Distillation paper~\cite{leal2021self} reports that RNN-T loss and soft target distillation loss together works well with weak teacher. We remain making weak teacher to strong teacher by knowledge distillation for future study. As the teacher is stronger, the distillation method should converge to the suggested method in this paper.
\section{Conclusion}
\label{sec:conclusion}

In this paper, we conducted extensive empirical studies for RNN-T distillation. Our results show that soft target distillation works better when both the teacher and student models have the same architecture. Otherwise, hard target distillation works better. We further demonstrate that soft target distillation achieved better results for self-training NST, where iterative distillation is applied to the same model. This enhances a $600$M Conformer RNN-T model to achieve a new SoTA WERs for LibriSpeech.

\section{ACKNOWLEDGMENTS}
\label{sec:ack}

\blind{
We thank Andrew Rosenberg, Bhuvana Ramabhadran, Bo Li, Chung-Cheng Chiu, CJ Zheng, Françoise Beaufays, Hasim Sak, Isabel Leal, James Qin, Jiahui Yu, Mohammadreza Ghodsi, Nikhil Siddhartha, Oren Litvin, Oscar Chang, Pedro Moreno Mengibar, Seungyeon Kim, Shefali Garg, Tara Sainath, Yonghui Wu, Zhehuai Chen, Zhouyuan Huo for helpful discussions.
}


\bibliographystyle{IEEEbib}
\newpage
\bibliography{ref}

\newpage 
\appendix
\section{Appendix}

\subsection{Training hyper parameters}
\label{sec:training_hyperparameters}

\subsubsection{Batch size}
\label{sec:batch_size}

We conduct both hard and soft target distillation with batch size $256$, $512$, $1024$, and $2048$. Hard target distillation has similar WERs over all the batch size. However, soft target distillation has better WERs with bigger batch size, as Tab~\ref{tab:batch} shows the average WER of all LibriSpeech testsets.

\begin{table}[htb]
    \centering
    \caption{LibriSpeech average WER comparison bewteen different batch size.} 
    \label{tab:batch}
    \begin{tabular}{lcccc}
        \toprule
        {\multirow{2}*{\textbf{Method}}}& \multicolumn{4}{c}{\textbf{Avg WER per batch size}} \\
         & {\textbf{256}} & {\textbf{512}} & {\textbf{1k}} & {\textbf{2k}} \\
        \midrule
        {soft target} & ${2.10}$  & ${2.05}$ & ${1.98}$  & $\textbf{1.95}$ \\
        {hard target} & ${2.00}$  & ${2.00}$ & ${1.98}$  & ${1.98}$ \\
        \bottomrule
    \end{tabular}
    %\vspace{-0.4cm}
\end{table}

\subsubsection{Learning hyper parameters}
\label{sec:learning_rate}

Hard target distillation is sensitive for learning hyper parameters such as learning rate and learning schedule. We use different learning rate and schedule for the encoder and the decoder, as Wav2vec 2.0 paper~\cite{baevski2020wav2vec} did, because the encoder is pre-trained. Otherwise, the training is easily diverged. On the other hands, soft target distillation is forgiving over wide range of learning rate and schedule. It allows us to have higher learning rate.

\subsection{SpecAugment}

To search the proper augmentation level, we multiply the number of mask count while fixing the max parameters. For example, 2x means that $4$ frequency masks and $20$ time masks, in Tab~\ref{tab:disc_aug}. As shown in Tab~\ref{tab:disc_aug}, $1x$ SpecAugment (\texttt{E1}) is better augmentation than heavier augmentations. More augmentations degrades the performance, as it causes too much mismatch between input features of teacher and student. 

\begin{table}[htb]
    \centering
    \caption{Soft distillation WERs comparison among different augmentations.} 
    \label{tab:spec_aug}
    \begin{tabular}{lcccc}
        \toprule
        {\multirow{2}*{\textbf{Augmentation}}}& \multicolumn{4}{c}{\textbf{WER}} \\
         & {\textbf{dev}} & {\textbf{dev-other}} & {\textbf{test}} & {\textbf{test-other}} \\
        \midrule
        {No Augment} & ${1.3}$  & ${2.5}$ & ${1.4}$  & ${2.6}$ \\
        \midrule
        {E12 (SpecAug)} & ${1.3}$  & ${2.5}$ & ${1.4}$  & ${2.6}$ \\
        {1.5x SpecAug} & ${1.4}$  & ${2.6}$ & ${1.4}$  & ${2.6}$ \\
        {2x SpecAug} & ${1.4}$  & ${2.7}$ & ${1.5}$  & ${2.7}$ \\
        {3x SpecAug} & ${1.4}$  & ${2.7}$ & ${1.6}$  & ${2.9}$ \\
        \bottomrule
    \end{tabular}
    %\vspace{-0.4cm}
\end{table}

\subsection{Consistency loss}
\label{sec:consistency_loss}

Consistency loss~\cite{hori2019cycle,xu2019data} (or Maximum Mean Discrepancy loss) is another natural choice of knowledge distillation for ASR model. We conduct Encoder-state-level consistency loss between teacher and student encoder outputs. We augment this loss as auxiliary loss to Eq~\ref{eq:overall_loss}, but it does not bring additional value.

\begin{table}[htb]
    \centering
    \caption{LibriSpeech WERs comparison with/without consistency loss.} 
    \label{tab:consistency}
    \begin{tabular}{lcccc}
        \toprule
        {\multirow{2}*{\textbf{Model}}}& \multicolumn{4}{c}{\textbf{WER}} \\
         & {\textbf{dev}} & {\textbf{dev-other}} & {\textbf{test}} & {\textbf{test-other}} \\
        \midrule
        {E12 (soft target)} & ${1.3}$  & ${2.5}$ & ${1.4}$  & ${2.6}$ \\
        {E12 + consistency} & ${1.3}$  & ${2.5}$ & ${1.4}$  & ${2.6}$ \\
        \bottomrule
    \end{tabular}
    %\vspace{-0.4cm}
\end{table}

\end{document}